\def\year{2019}\relax
\documentclass[letterpaper]{article} 
\usepackage{aaai19}  
\usepackage{times}  
\usepackage{helvet}  
\usepackage{courier}  
\usepackage{url}  
\usepackage{graphicx}  

\usepackage{amsmath,amssymb,amsfonts}
\usepackage{algorithmic}
\usepackage{graphicx}
\usepackage{tabularx}
\usepackage{textcomp}
\usepackage{tikz}
\usepackage{float}
\usepackage{pgfplots}
\usepackage{booktabs}

\usetikzlibrary{calc}

\begin{document}
%
\title{Ensembles of Nested Dichotomies with Multiple Subset Evaluation}
\author{Tim Leathart, Eibe Frank, Bernhard Pfahringer and Geoffrey Holmes\\
Department of Computer Science, University of Waikato, New Zealand}
\maketitle
\begin{abstract}
A system of nested dichotomies is a method of decomposing a multi-class problem into a collection of binary problems. Such a system recursively applies binary splits to divide the set of classes into two subsets, and trains a binary classifier for each split. Many methods have been proposed to perform this split, each with various advantages and disadvantages. In this paper, we present a simple, general method for improving the predictive performance of nested dichotomies produced by any subset selection techniques that employ randomness to construct the subsets. We provide a theoretical expectation for performance improvements, as well as empirical results showing that our method improves the root mean squared error of nested dichotomies, regardless of whether they are employed as an individual model or in an ensemble setting.	
\end{abstract}

\section{Introduction}
Multi-class classification problems are commonplace in real world applications. Some models, like neural networks and random forests, are inherently able to operate on multi-class data directly, while other models, such as classic support vector machines, can only be used for binary (two-class) problems. The standard way to bypass this limitation is to convert the multi-class classification problem into a series of binary problems. There exist several methods of performing this decomposition, the most well-known including one-vs-rest~\cite{rifkin2004defense}, pairwise classification~\cite{hastie1998classification} and error-correcting output codes~\cite{dietterich1995solving}. Models that are directly capable of working with multi-class problems may also see improved accuracy from such a decomposition~\cite{mayoraz1997decomposition,furnkranz2002round,pimenta2005study}.

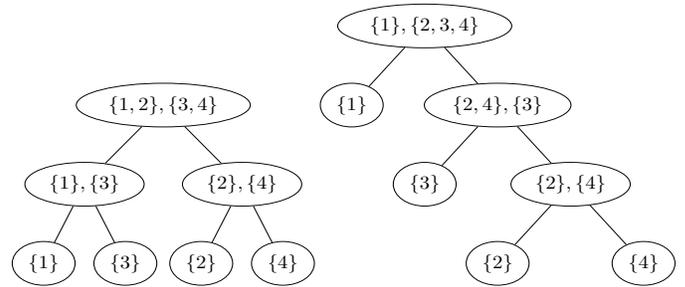
\begin{figure}[t!]
	\centering
	\resizebox{0.5\textwidth}{3.75cm}{%
		\begin{tikzpicture}
			\tikzstyle{level 1}=[sibling distance = 27mm]
			\tikzstyle{level 2}=[sibling distance = 14mm]
			\usetikzlibrary{shapes}
			\node[ellipse,draw](z){$\{1,2\},\{3,4\}$}
			  child
			  {
			  	node[ellipse,draw]{$\{1\},\{3\}$}
			  	child
			  	{
			  		node[ellipse,draw]{$\{1\}$}
			  	}
			  	child
			  	{
			  		node[ellipse,draw]{$\{3\}$}
			  	}
			  }
			  child
			  {
		    		node[ellipse,draw]{$\{2\},\{4\}$} 
		    		child
		    		{
		    			node[ellipse,draw] {$\{2\}$}
		    		} 
		    		child
		    		{
		    			node[ellipse,draw] {$\{4\}$}
		    		}
		    	  };
		\end{tikzpicture}
		\begin{tikzpicture}
			\tikzstyle{level 1}=[sibling distance = 25mm]
			\usetikzlibrary{shapes}
			\node[ellipse,draw](z){$\{1\},\{2,3,4\}$}
			  child
			  {
			  	node[ellipse,draw]{$\{1\}$}
			  }
			  child
			  {
		    		node[ellipse,draw]{$\{2,4\},\{3\}$} 
		    		child
		    		{
		    			node[ellipse,draw] {$\{3\}$}
		    		} 
		    		child
		    		{
		    			node[ellipse,draw] {$\{2\},\{4\}$}
		    			child
		    			{
		    				node[ellipse,draw] {$\{2\}$}
		    			}
		    			child
		    			{
		    				node[ellipse,draw] {$\{4\}$}
		    			}
		    		}
		    	  };
		\end{tikzpicture}
	}
	\caption{\label{fig:nd_example} Two examples of nested dichotomies for a four class problem. Each node shows the subset of classes used to train the internal binary classifiers.}
\end{figure}

The use of ensembles of nested dichotomies is one such method for decomposing a multi-class problem into several binary problems. It has been shown to outperform one-vs-rest and perform competitively compared to the aforementioned methods~\cite{frank2004ensembles}. In a nested dichotomy~\cite{fox1997applied}, the set of classes is recursively split into two subsets in a tree structure. Two examples of nested dichotomies for a four class problem are shown in Figure~\ref{fig:nd_example}. At each split node of the tree, a binary classifier is trained to discriminate between the two subsets of classes. Each leaf node of the tree corresponds to a particular class. To obtain probability estimates for a particular class from a nested dichotomy, assuming the base learner can produce probability estimates, one can simply compute the product of the binary probability estimates along the path from the root node to the leaf node corresponding to the class. 

For non-trivial multi-class problems, the space of potential nested dichotomies is very large.  An ensemble classifier can be formed by choosing suitable decompositions from this space. In the original formulation of ensembles of nested dichotomies, decompositions are sampled with uniform probability~\cite{frank2004ensembles}, but several other more sophisticated methods for splitting the set of classes have been proposed~\cite{dong2005ensembles,duarte2012nested,leathart2016building}. Superior performance is achieved when ensembles of nested dichotomies are trained using common ensemble learning methods like bagging or boosting~\cite{rodriguez2010forests}. 

In this paper, we describe a simple method that can improve the predictive performance of nested dichotomies by considering several splits at each internal node. Our technique can be applied to nested dichotomies built with almost any subset selection method, only contributing a constant factor to the training time and no additional cost when obtaining predictions. It has a single hyperparameter $\lambda$ that gives a trade-off between predictive performance and training time, making it easy to tune for a given learning problem. It is also very easy to implement.

The paper is structured as follows. First, we describe existing methods for class subset selection in nested dichotomies. Following this, we describe our method and provide a theoretical expectation of performance improvements. We then discuss related work, before presenting and discussing empirical results for our experiments. Finally, we conclude and discuss future research directions.

\section{Class Subset Selection Methods\label{sec:subset_selection_methods}}
At each internal node $i$ of a nested dichotomy, the set of classes present at the node $\mathcal{C}_i$ is split into two non-empty, non-overlapping subsets, $\mathcal{C}_{i1}$ and $\mathcal{C}_{i2}$. In this section, we give an overview of existing class subset selection methods for nested dichotomies. Note that other methods than those listed here have been proposed for constructing nested dichotomies---these are not suitable for use with our method and are discussed later in Related Work.

\pgfplotscreateplotcyclelist{growthlist}{%
black, solid, every mark/.append style={solid, fill=black}, mark=*\\%
red, solid, every mark/.append style={solid, fill=red}, mark=square*\\%
brown, solid, every mark/.append style={solid, fill=brown}, mark=x\\%
blue, solid, every mark/.append style={solid, fill=blue}, mark=triangle*\\%
red, dashed, every mark/.append style={solid, fill=white},mark=square*\\%
brown, dashed, every mark/.append style={solid, fill=white},mark=x\\%
blue, dashed, every mark/.append style={solid, fill=white},mark=triangle*\\%
}

\pgfplotstableread[col sep=tab,header=true]{
n	cbnd	nd	rpnd	clustering
2	1	1	1	1
3	3	3	1	1
4	3	15	5	1
5	30	105	15	1
6	90	945	36	1
7	315	10395	182	1
8	315	135135	470	1
9	11340	2027025	1254	1
10	113400	34459425	7002	1
11	1247400	654729075	28189	1
12	3742200	13749310575	81451	1
}\data

\begin{figure}[t!]
	\centering
	    \resizebox{0.43\textwidth}{0.4\textwidth}{%
		    \begin{tikzpicture}     
		        \begin{axis}[    
				        cycle list name=growthlist,
			            width=7.5cm,
			            height=7.5cm,
			            line width=0.5pt,   
			            mark size=1.5pt,    
			            ymin=0,
			            ymode=log,
			            ymax=1e11,
			            ytick={1e2,1e4,1e6,1e8,1e10},
			            xtick={2,4,6,8,10,12},
			            log basis y=10, 
			            xmin=2,
			            xmax=12,
			            grid=major,
			            ylabel={Size of sample space},
			            xlabel={Number of classes},
			            enlarge y limits={abs=0},
				        legend cell align=left,
			            legend style={at={(0.025,0.975)},anchor=north west}
		            ]       
			        \addplot table [y=nd, x=n] {\data};   
			        \addplot table [y=cbnd, x=n] {\data};   
			        \addplot table [y=rpnd, x=n] {\data};  
			        \legend{Random Selection\\Class Balanced\\Random-Pair (estimated)\\}
		        \end{axis}
		    \end{tikzpicture}
		}
	\caption{\label{fig:growth} Growth functions for each subset selection method discussed.}
\end{figure}
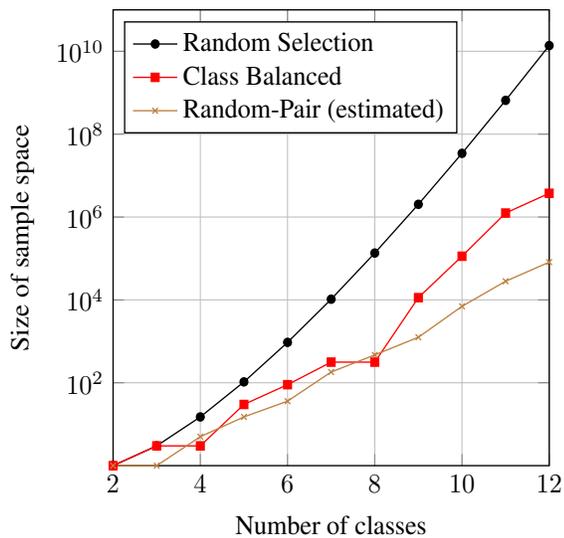

\usepgfplotslibrary{fillbetween}

\pgfplotstableread[col sep=tab,header=true]{
n	cbnd	nd	rpnd	clustering
2	1	1	1	1
3	3	3	1	1
4	3	15	5	1
5	30	105	15	1
6	90	945	36	1
7	315	10395	182	1
8	315	135135	470	1
9	11340	2027025	1254	1
10	113400	34459425	7002	1
11	1247400	654729075	28189	1
12	3742200	13749310575	81451	1
}\data

\pgfplotstableread[col sep=tab,header=true]{
n	1	3f	3l	5f	5l	7f	7l	9f	9l
2	1	1	1	1	1	1	1	1	1	
3	3	1	1	1	1	1	1	1	1	
4	15	5	5	3	3	1	1	1	1	
5	105	25	33	13	21	9	9	7	7	
6	945	185	281	81	177	25	73	23	59	
7	10395	1625	2825	621	1773	249	729	175	587	
8	135135	17205	33141	5795	20595	1737	8409	1071	6767
9	2027025	210225	444033	63049	271737	17409	109137	10185	87817	
10	34459425	2924025	6700761	785913	4022217	193137	1582305	100569	1271297	
11	654729075	45535425	112525281	11026537	66045753	2395905	25335537	1170729	20311969	
12	13749310575	785158725	2082421093	171983907	1192218291	33875889	444257505	15305913	355153337	
}\randomdata

\pgfplotstableread[col sep=tab,header=true]{
n	1	3	5	7	9
2	1	1	1	1	1	
3	3	1	1	1	1	
4	3	1	1	1	1	
5	30	8	6	4	2	
6	90	8	6	4	2	
7	315	33	31	29	27	
8	315	33	31	29	27	
9	11340	992	732	480	236	
10	113400	7936	4392	1920	472	
11	1247400	29440	16488	7296	1816	
12	3742200	29440	16488	7296	1816	
}\cbdata

\pgfplotstableread[col sep=tab,header=true]{
n	1	3	5	7	9	
2	1	1	1	1	1	
3	1	1	1	1	1	
4	4	1	1	1	1	
5	15	2	1	1	1	
6	37	5	3	1	1	
7	192	20	7	5	3	
8	511	50	19	9	7	
9	1390	104	59	23	12	
10	7989	335	129	77	33	
11	33241	856	240	170	108	
12	99174	1911	712	315	232
}\rpdata

\begin{figure*}[t!]
    \centering
    {
    	\begin{tabular}{cccc}
		    \resizebox{0.32\textwidth}{!}{%
			    \begin{tikzpicture}     
			        \begin{axis}[    
					        cycle list name=growthlist,
				            width=7.5cm,
				            height=7.5cm,
				            line width=0.5pt, 
				            mark size=1.5pt,                                  
				            ymin=0,
				            ymode=log,
				            ymax=1e11,
				            ytick={1e2,1e4,1e6,1e8,1e10},
				            xtick={2,4,6,8,10,12},
				            log basis y=10, 
				            xmin=2,
				            xmax=12,
				            grid=major,
				            ylabel={Size of sample space},
				            xlabel={Number of classes},
				            title={Random Selection},
				            enlarge y limits={abs=0},
					        legend cell align=left,
				            legend style={at={(0.025,0.975)},anchor=north west}
			            ]       
				        \addplot table [y=1, x=n] {\randomdata};   
				        \addplot table [name path=A, y=3l, x=n] {\randomdata};   
				        \addplot table [name path=5l, y=5l, x=n] {\randomdata};  
				        \addplot table [name path=7l, y=7l, x=n] {\randomdata};
				       	\addplot table [name path=B, y=3f, x=n] {\randomdata};   
				        \addplot table [name path=5f, y=5f, x=n] {\randomdata};  
				        \addplot table [name path=7f, y=7f, x=n] {\randomdata};
						\legend{$\lambda=1$\\$\lambda=3$\\$\lambda=5$\\$\lambda=7$\\}
			        \end{axis}
			    \end{tikzpicture}
			    
			} & 
			\resizebox{0.3\textwidth}{!}{%
			    \begin{tikzpicture}     
			        \begin{axis}[    
					        cycle list name=growthlist,
				            width=7.5cm,
				            height=7.5cm,
				            line width=0.5pt,  
    			            mark size=1.5pt,                                                         
				            ymin=0,
				            ymode=log,
				            ymax=1e11,
				            ytick={1e2,1e4,1e6,1e8,1e10},
				            xtick={2,4,6,8,10,12},
				            log basis y=10, 
				            xmin=2,
				            xmax=12,
				            grid=major,
				            xlabel={Number of classes},
				            title={Class Balanced Selection},
				            enlarge y limits={abs=0},
					        legend cell align=left,
				            legend style={at={(0.025,0.975)},anchor=north west}
			            ]       
				        \addplot table [y=1, x=n] {\cbdata};   
				        \addplot table [name path=A, y=3, x=n] {\cbdata};   
				        \addplot table [name path=5l, y=5, x=n] {\cbdata};  
				        \addplot table [name path=7l, y=7, x=n] {\cbdata};
				        
				        \legend{$\lambda=1$\\$\lambda=3$\\$\lambda=5$\\$\lambda=7$\\}
			        \end{axis}
			    \end{tikzpicture}
			    
			} & 
			\resizebox{0.3\textwidth}{!}{%
			    \begin{tikzpicture}     
			        \begin{axis}[    
					        cycle list name=growthlist,
				            width=7.5cm,
				            height=7.5cm,
				            line width=0.5pt, 
				            mark size=1.5pt,                                  
				            ymin=0,
				            ymode=log,
				            ymax=1e11,
				            ytick={1e2,1e4,1e6,1e8,1e10},
				            xtick={2,4,6,8,10,12},
				            log basis y=10, 
				            xmin=2,
				            xmax=12,
				            grid=major,
				            xlabel={Number of classes},
				            title={Random-Pair Selection},
				            enlarge y limits={abs=0},
					        legend cell align=left,
				            legend style={at={(0.025,0.975)},anchor=north west}
			            ]       
				        \addplot table [y=1, x=n] {\rpdata};   
				        \addplot table [name path=A, y=3, x=n] {\rpdata};   
				        \addplot table [name path=5l, y=5, x=n] {\rpdata};  
				        \addplot table [name path=7l, y=7, x=n] {\rpdata};
				        
				        \legend{$\lambda=1$\\$\lambda=3$\\$\lambda=5$\\$\lambda=7$\\}
			        \end{axis}
			    \end{tikzpicture}   
			} & \\
		\end{tabular}
	}
\caption{\label{fig:growthrate} Left: Growth functions for random selection, with multiple subset evaluation and $\lambda \in \{1,3,5,7\}$. Solid lines indicate the upper bound, and dotted lines indicate the lower bound. Middle: Considering class-balanced selection instead of random selection. Right: Growth functions for random-pair selection.}
\end{figure*}
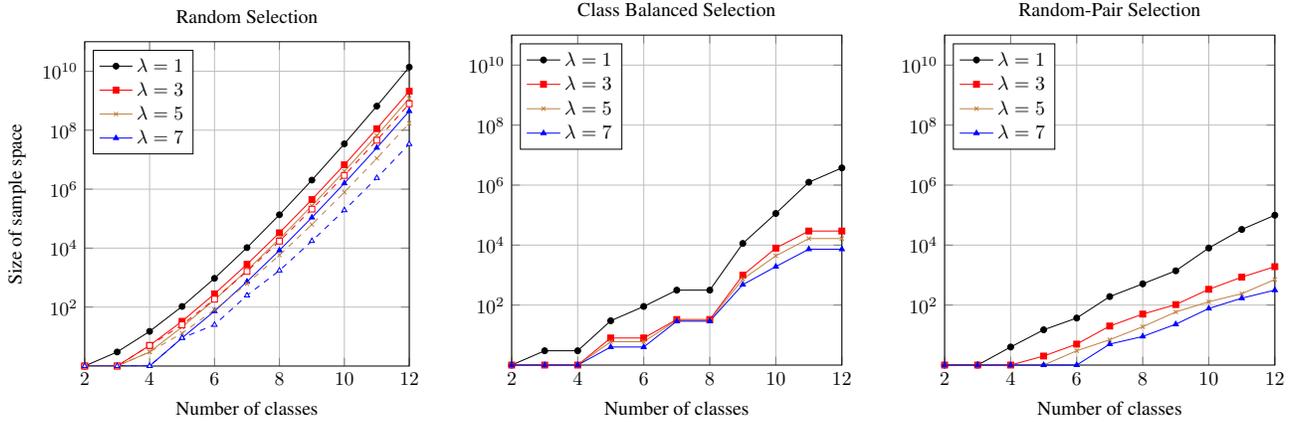

\subsection{Random Selection}
The most basic form of class subset selection method, originally proposed in~\cite{frank2004ensembles}, is to split the set of classes into two subsets such that each member of the space of nested dichotomies has an equal probability of being sampled. This approach has several attractive qualities. It is simple to compute, and does not scale with the size of the dataset, making it suitable for datasets of any size. Furthermore, for an $n$-class problem, the number of possible nested dichotomies is very large, given by the recurrence relation
\begin{align*}
	T(n) = (2n-3) \times T(n-1)
\end{align*}

where $T(1) = 1$. This ensures that, in an ensemble of nested dichotomies, there is a high level of diversity amongst ensemble members. We refer to this function that relates the number of classes to the size of the sample space of nested dichotomies for a given subset selection method as the \textit{growth function}. Growth functions for each method discussed in this section are compared in Figure~\ref{fig:growth}.

\subsection{Balanced Selection}
An issue with random selection is that it can produce very unbalanced tree structures. While the number of internal nodes (and therefore, binary models) is the same in any nested dichotomy for the same number of classes, an unbalanced tree often implies that the internal binary models are trained on large datasets near the leaves, which has a negative effect on the time taken to train the full model. Deeper subtrees also provide more opportunity for estimation errors to accumulate. Dong~\textit{et. al.} mitigate this effect by enforcing $\mathcal{C}_i$ to be split into two subsets $\mathcal{C}_{i1}$ and $\mathcal{C}_{i2}$ such that ${abs}(|\mathcal{C}_{i1}| - |\mathcal{C}_{i2}|) \leq 1$~\cite{dong2005ensembles}. This has been shown empirically to have little effect on the accuracy in most cases, while reducing the time taken to train nested dichotomies. Balanced selection has greater benefits for problems with many classes. 

It is clear that the sample space of random nested dichotomies is larger than that of class balanced nested dichotomies, but it is still large enough to ensure sufficient ensemble diversity. The growth function for class balanced nested dichotomies is given by
\begin{align*}
	T_{CB}(n) = 
	\begin{cases}
		\frac{1}{2} \binom{n}{n/2} T_{CB}(\frac{n}{2}) T_{CB}(\frac{n}{2}), & \text{if } n \text{ is even} \\
		\binom {n}{(n+1)/2} T_{CB}(\frac{n+1}{2}) T_{CB}(\frac{n-1}{2}), & \text{if } n \text{ is odd} \\
	\end{cases}
\end{align*}

where $T_{CB}(2) = T_{CB}(1) = 1$~\cite{dong2005ensembles}. 

Dong~\textit{et. al.} also explored a form of balancing where the amount of data in each subset is roughly equal, which gave similar results for datasets with unbalanced classes~\cite{dong2005ensembles}.

\subsection{Random-Pair Selection}
Random-pair selection provides a non-deterministic method of creating $\mathcal{C}_{i1}$ and $\mathcal{C}_{i2}$ that groups similar classes together~\cite{leathart2016building}. In random-pair selection, the base classifier is used directly to identify similar classes in $\mathcal{C}_i$. First, a random pair of classes $c_1, c_2 \in \mathcal{C}_i$ is selected, and a binary classifier is trained on just these two classes. Then, the remaining classes are classified with this classifier, and its predictions are stored as a confusion matrix $M$. $\mathcal{C}_{i1}$ and $\mathcal{C}_{i2}$ are constructed by
\begin{align*}
	\mathcal{C}_{i1} &= \{ c \in \mathcal{C}_i \setminus \{c_1, c_2\} : M_{c, c_1} \leq M_{c, c_2} \} \cup \{c_1\} \\
	\mathcal{C}_{i2} &= \{ c \in \mathcal{C}_i \setminus \{c_1, c_2\} : M_{c, c_1} > M_{c, c_2} \} \cup \{c_2\}
\end{align*}
where $M_{i,j}$ is defined as the number of examples of class $j$ that were classified as class $i$ by the binary classifier. In other words, a class is assigned to $\mathcal{C}_{i1}$ if it is less frequently confused with $c_1$ than with $c_2$, and to $\mathcal{C}_{i2}$ otherwise. Finally, the binary classifier is re-trained on the new meta-classes $\mathcal{C}_{i1}$ and $\mathcal{C}_{i2}$. This way, each binary split is more easily separable for the base learner than a completely random split, but also exhibits a degree of randomness, which leads to diverse and high-performing ensembles.

Due to the fact that the size of the sample space of nested dichotomies under random-pair selection is dependent on the dataset and base learner (different initial random pairs may lead to the same split), it is not possible to provide an exact expression for the growth function $T_{RP}(n)$; using logistic regression as the base learner~\cite{leathart2016building}, it has been empirically estimated to be 
\begin{align*}
	T_{RP}(n) = p(n)T_{RP}(\frac{n}{3})T_{RP}(\frac{2n}{3}) 	\label{eqn:random_pair_estimation}
\end{align*}
where
\begin{align*}
	p(n) = 0.3812n^2 - 1.4979n + 2.9027
\end{align*}				
and $T_{RP}(2) = T_{RP}(1) = 1$.

\section{Multiple Subset Evaluation\label{sec:multiple_subset_selection}}

In existing class subset selection methods, at each internal node $i$, a single class split $(\mathcal{C}_{i1}, \mathcal{C}_{i2})$ of $\mathcal{C}_i$ is considered, produced by some splitting function $S(\mathcal{C}_i) : \mathbb{N}^n \rightarrow \mathbb{N}^a \times \mathbb{N}^b$ where $a+b=n$. Our approach for improving the predictive power of nested dichotomies is a simple extension. We propose to, at each internal node $i$, consider $\lambda$ subsets $\{(\mathcal{C}_{i1}, \mathcal{C}_{i2})_1 \dots (\mathcal{C}_{i1}, \mathcal{C}_{i2})_\lambda\}$ and choose the split for which the corresponding model has the lowest training root mean squared error (RMSE). The RMSE is defined as the square root of the Brier score~\cite{brier1950verification} divided by the number of classes:
\begin{displaymath}
	\textrm{RMSE} = \sqrt{\frac{1}{nm}\sum_{i=1}^n \sum_{j=1}^m (\hat{y}_{ij} - y_{ij})^2 }
\end{displaymath}  
where $n$ is the number of instances, $m$ is the number of classes, $\hat{y}_{ij}$ is the estimated probability that instance $i$ is of class $j$, and $y_{ij}$ is $1$ if instance $i$ actually belongs to class $j$, and $0$ otherwise. RMSE is chosen over other measures such as classification accuracy because it is smoother and a better indicator of generalisation performance. Previously proposed methods with single subset selection can be considered a special case of this method where $\lambda = 1$. 
 
 Although conceptually simple, this method has several attractive qualities, which are now discussed. 
\paragraph{Predictive Performance.} It is clear that by choosing the best of a series of models at each internal node, the overall performance should improve, assuming the size of the sample space of nested dichotomies is not hindered to the point where ensemble diversity begins to suffer.

\paragraph{Generality.} Multiple subset evaluation is widely applicable. If a subset selection method $S$ has some level of randomness, then multiple subset evaluation can be used to improve the performance. One nice feature is that advantages pertaining to $S$ are retained. For example, if class-balanced selection is chosen due to a learning problem with a very high number of classes, we can boost the predictive performance of the ensemble while keeping each nested dichotomy in the ensemble balanced. If random-pair selection is chosen because the computational budget for training is high, then we can improve the predictive performance further than single subset selection in conjunction with random-pair selection.

\paragraph{Simplicity.} Implementing multiple subset evaluation is very simple. Furthermore, the computational cost for evaluating multiple subsets of classes scales linearly in the size of the tuneable hyperparameter $\lambda$, making the tradeoff between predictive performance and training time easy to navigate. Additionally, multiple subset evaluation has no effect on prediction times.

Higher values of $\lambda$ give diminishing returns on predictive performance, so a value that is suitable for the computational budget should be chosen. When training an ensemble of nested dichotomies, it may be desirable to adopt a \textit{class threshold}, where single subset selection is used if fewer than a certain number of classes is present at an internal node. This reduces the probability that the same subtrees will appear in many ensemble members, and therefore reduce ensemble diversity. In the lower levels of the tree, the number of possible binary problems is relatively low~(Fig.~\ref{fig:growthrate}).

\subsection{\label{sec:growth_functions_effect}Effect on Growth Functions}
Performance of an ensemble of nested dichotomies relies on the size of the sample space of nested dichotomies, given an $n$-class problem, to be relatively large. Multiple subset evaluation removes the $\lambda-1$ class splits that correspond to the worst-performing binary models at each internal node $i$ from being able to be used in the tree. The effect of multiple subset evaluation on the growth function is non-deterministic for random selection, as the sizes of $\mathcal{C}_{i1}$ and $\mathcal{C}_{i2}$ affect the values of the growth function for the subtrees that are children of $i$. The upper bound occurs when all worst-performing splits isolate a single class, and the lower bound is given when all worst-performing splits are class-balanced. Class-balanced selection, on the other hand, is affected deterministically as the size of $\mathcal{C}_{i1}$ and $\mathcal{C}_{i2}$ are the same for the same number of classes. 

Growth functions for values of $\lambda \in \{1, 3, 5, 7\}$, for random, class balanced and random-pair selection methods, are plotted in Figure~\ref{fig:growthrate}. The growth curves for random and class balanced selection were generated using brute-force computational enumeration, while the effect on random-pair selection is estimated.

\begin{figure*}[t!]
	\centerline{
		\includegraphics[width=0.95\textwidth]{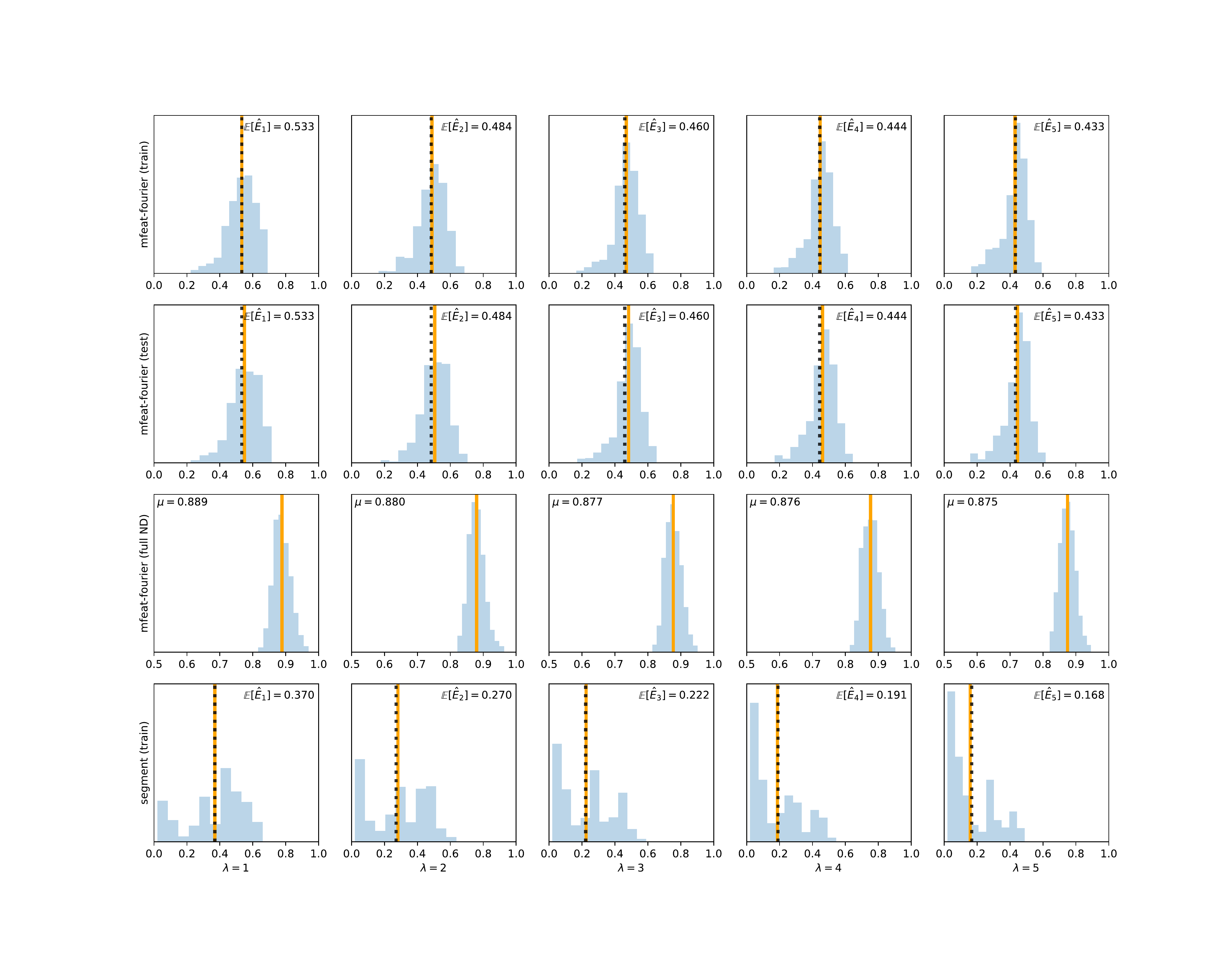} 
	}
	\caption{\label{fig:norm_drawn} Empirical distribution of RMSE of logistic regression trained on random binary class splits, for values of $\lambda$ from one to five. Shaded region indicates empirical histogram, orange vertical line shows the empirical mean, and the black dotted vertical line is expected value, estimated from (\ref{eqn:expected_order_statistics}). Top two rows: train and test RMSE of logistic regression trained on random binary class splits of \texttt{mfeat-fourier} UCI dataset. For the test data, the approximated value of $\mathbb{E}[E_\lambda]$ is estimated from the mean and standard deviation of the train error. Third row: train RMSE of a nested dichotomy built with random splits and multiple-subset evaluation, trained on \texttt{mfeat-fourier} for different values of $\lambda$. Bottom row: train RMSE of logistic regression trained on random binary class splits of \texttt{segment} data.}
\end{figure*}

\subsection{Analysis of error\label{sec:theoretical}}
In this section, we provide a theoretical analysis showing that performance of each internal binary model is likely to be improved by adopting multiple subset evaluation. We also show empirically that the estimates of performance improvements are accurate, even when the assumptions are violated.

 Let $E$ be a random variable for the training root mean squared error (RMSE) for some classifier for a given pair of class subsets $\mathcal{C}_{i1}$ and $\mathcal{C}_{i2}$, and assume $E \sim N(\mu, \sigma^2)$ for a given dataset under some class subset selection scheme. 
For a given set of $\lambda$ selections of subsets $\mathcal{S} = \{(\mathcal{C}_{i1}, \mathcal{C}_{i2})_1, \dots, (\mathcal{C}_{i1}, \mathcal{C}_{i2})_\lambda\}$ and corresponding training \mbox{RMSE}s $\mathcal{E} = \{E_1, \dots, E_\lambda\}$, let $\hat{E}_\lambda = min(\mathcal{E})$. There is no closed form expression for the expected value of $\hat{E}_\lambda$, the minimum of a set of normally distributed random variables, but an approximation is given by 
\begin{equation}
	\mathbb{E}[\hat{E}_\lambda] \approx \mu + \sigma \Phi^{-1} \Bigg( \frac{1-\alpha}{\lambda-2\alpha + 1}\Bigg) \label{eqn:expected_order_statistics}
\end{equation}
where $\Phi^{-1}(x)$ is the inverse normal cumulative distribution function~\cite{royston1982algorithm}, and the \textit{compromise value} $\alpha$ is the suggested value for $\lambda$ given by Harter~(\citeyear{harter1961expected}).\footnote{Appropriate values for $\alpha$ for a given $\lambda$ can be found in Table 3 of~\cite{harter1961expected}.}  

Figure~\ref{fig:norm_drawn} illustrates how this expected value changes when increasing values of $\lambda$ from $1$ to $5$. The first two rows show the distribution of $E$ and estimated $\mathbb{E}[\hat{E}_\lambda]$ on the UCI dataset \texttt{mfeat-fourier}, for a logistic regression model trained on 1,000 random splits of the class set $\mathcal{C}$. These rows show the training and testing RMSE respectively, using 90\% of the data for training and the rest for testing. Note that as $\lambda$ increases, the distribution of the train and test error shifts to lower values and the variance decreases.
 
 This reduction in error affects each binary model in the tree structure, so the effects accumulate when constructing a nested dichotomy. The third row shows the distribution of RMSE of 1,000 nested dichotomies trained with multiple subset evaluation on \texttt{mfeat-fourier}, using logistic regression as the base learner, considering increasing values of $\lambda$. As expected, a reduction in error with diminishing returns is seen as $\lambda$ increases.
 
  In order to show an example of how the estimate from~(\ref{eqn:expected_order_statistics}) behaves when the error is not normally distributed, the distribution of $E$ for logistic regression trained on the \texttt{segment} UCI data is plotted in the bottom row. This assumption is commonly violated in real datasets, as the distribution is often skewed towards zero error. As with the other examples, 1,000 different random choices for $\mathcal{C}_1$ and $\mathcal{C}_2$ were used to generate the histogram. Although the distribution in this case is not very well modelled by a Gaussian, the approximation of $\mathbb{E}[\hat{E}_\lambda]$ from~(\ref{eqn:expected_order_statistics}) still closely matches the empirical mean. This shows that even when the normality assumption is violated, performance gains of the same degree can be expected. This example is not cherry picked; the same behaviour was observed on the entire collection of datasets used in this study.

\section{Related Work\label{sec:related_work}}
Splitting a multi-class problem into several binary problems in a tree structure is a general technique that has been referred to by different names in the literature. For example, in a multi-class classification context, nested dichotomies in the broadest sense of the term have been examined as filter trees, conditional probability trees, and label trees. \citeauthor{beygelzimer2009conditional} proposed algorithms which build balanced trees and demonstrate the performance on datasets with very large numbers of classes. Filter trees, with deterministic splits~\cite{beygelzimer2009error}, as well as conditional probability trees, with probabilistic splits~\cite{beygelzimer2009conditional}, were explored. \citeauthor{bengio2010label} (\citeyear{bengio2010label}) define a tree structure and optimise all internal classifiers simultaneously to minimise the tree loss. They also propose to learn a low-dimensional embedding of the labels to improve performance, especially when a very large number of classes is present. \citeauthor{melnikov2018effectiveness} (\citeyear{melnikov2018effectiveness}) also showed that a method called bag-of-$k$ models---simply sampling $k$ random nested dichotomies and choosing the best one based on validation error---gives competitive predictive performance to the splitting heuristics discussed so far for individual nested dichotomies (i.e., not trained in an ensemble). However, it is very expensive at training time, as $k$ independent nested dichotomies must be constructed and tested on a validation set.

A commonality of these techniques is that they attempt to build a single nested dichotomy structure with the best performance. Nested dichotomies that we consider in this paper, while conceptually similar, differ from these methods because they are intended to be trained in an ensemble setting, and as such, each individual nested dichotomy is not built with optimal performance in mind. Instead, a group of nested dichotomies is built to maximise ensemble performance, so diversity amongst the ensemble members is key~\cite{kuncheva2003measures}. 

Nested dichotomies based on clustering~\cite{duarte2012nested}, are deterministic and used in an ensemble by resampling or reweighting the input. They are built by finding the two classes in $\mathcal{C}_i$ for which the class centroids are furthest from each other by some distance metric. The remainder of the classes are grouped based on the distance of their centroids from the initial two centroids.

\citeauthor{wever2018ensembles} (\citeyear{wever2018ensembles}) utilise genetic algorithms to build nested dichotomies. In their method, a population of random nested dichotomies is sampled and runs through a genetic algorithm for several generations. The final nested dichotomy is chosen as the best performing model on a held-out validation set. An ensemble of $k$ nested dichotomies is produced by initialising $k$ individual populations, independently evolving each population, and taking the best-performing model from each population.

\section{Experimental Results\label{sec:experiments}}
All experiments were conducted in WEKA 3.9~\cite{hall2009weka}, and performed with 10 times 10-fold cross validation. We use class-balanced nested dichotomies and nested dichotomies built with random-pair selection and logistic regression as the base learner. 

For both splitting methods, we compare values of $\lambda \in \{1,3,5,7\}$ in a single nested dichotomy structure, as well as in ensemble settings with bagging~\cite{breiman1996bagging} and AdaBoost~\cite{freund1996game}. The default settings in WEKA were used for the \texttt{Logistic} classifier as well as for the \texttt{Bagging} and \texttt{AdaBoostM1} meta-classifiers. We evaluate performance on a collection of datasets taken from the UCI repository~\cite{lichman2013uci}, as well as the MNIST digit recognition dataset~\cite{lecun1998gradient}. Note that for MNIST, we report results of 10-fold cross-validation over the entire dataset rather than the usual train/test split. Datasets used in our experiments, and their number of classes, instances and features, are listed in Table~\ref{tab:datasets}.

We provide critical difference plots~\cite{demvsar2006statistical} to summarise the results of the experiments. These plots present average ranks of models trained with differing values of $\lambda$. Models producing results that are not significantly different from each other at the 0.05 significance level are connected with a horizontal black bar. Full results tables showing RMSE for each experimental run, including significance tests, are available in the supplementary materials.

\subsection{Individual Nested Dichotomies}
\begin{table}[t]
	\centering
	\caption{\label{tab:datasets}The datasets used in our experiments.}
	
	\begin{tabular}{lccc}
		\toprule
		\textbf{Dataset} & \textbf{Classes} & \textbf{Instances} & \textbf{Features} \\
		\midrule
		audiology & 24 & 226 & 70\\
		krkopt & 18 & 28056 & 7\\
		LED24 & 10 & 5000 & 25\\
		letter & 26 & 20000 & 17\\
		mfeat-factors & 10 & 2000 & 217\\
		mfeat-fourier & 10 & 2000 & 77\\
		mfeat-karhunen & 10 & 2000 & 65 \\
		mfeat-morph & 10 & 2000 & 7 \\
		mfeat-pixel & 10 & 2000 & 241\\
		MNIST & 10 & 70000 & 784\\
		optdigits & 10 & 5620 & 65\\
		page-blocks & 5 & 5473 & 11  \\
		pendigits & 10 & 10992 & 17\\
		segment & 7 & 2310 & 20\\
		usps & 10 & 9298 & 257\\
		vowel & 11 & 990 & 14\\
		yeast & 10 & 1484 & 9\\
		\bottomrule	
	\end{tabular}
\end{table}

Restricting the sample space of nested dichotomies through multiple subset evaluation is expected to have a greater performance impact on smaller ensembles than larger ones. This is because in a larger ensemble, a poorly performing ensemble member does not have a large impact on the overall performance. On the other hand, in a small ensemble, one poorly performing ensemble member can degrade the ensemble performance significantly. In the extreme case, where a single nested dichotomy is trained, there is no need for ensemble diversity, so a technique for improving the predictive performance of an individual nested dichotomy should be effective. Therefore, we first compare the performance of single nested dichotomies for different values of~$\lambda$. 

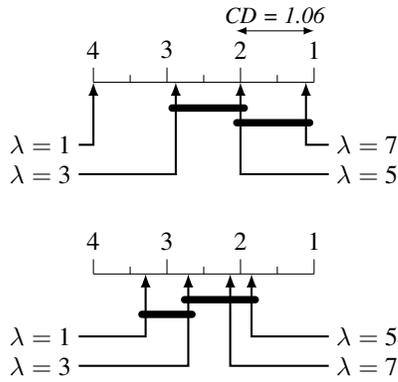
\begin{figure}[t]
    \centering
	    \resizebox{0.3\textwidth}{!}
	    {
	        \begin{tikzpicture}[>=latex]
	
	            \def\nemnum{4}        
	            \def\signum{2}        
	            \def\s{1*}                
	            \def\cd{1.06}        
	
	            \coordinate (cmin) at (\s1,0);
	            \coordinate (cmax) at (\s\nemnum,0);
		        \coordinate (cleft) at ($(cmin)-(.2,(\signum*.25-.05)$);
		        \coordinate (cright) at ($(cmax)+(.2,-(\signum*.25-.05)$);
		        \coordinate (csig) at (0,-.15);
		    
		        \draw (cmin) -- (cmax);
		    
		        \foreach \c in {1,2,...,\nemnum}{
		            \draw ($(cmin)+(cmax)-(\s\c,0)$) -- ($(cmin)+(cmax)-(\s\c,-.2)$) node[above] {\c};
		             \ifthenelse{\c < \nemnum} {
		                \draw ($(\s\c,0)+(\s.5,0)$) -- ($(\s\c,0)+(\s.5,.1)$);
		              }
		        }
		    
		        \draw[<->] ($(cmax)+(0,.7)$) -- node[above=.10cm,inner sep=0,outer sep=0] {\small{\textit{CD = 1.06}}} ($(cmax)-(\s\cd,-.7)$);
		      
		        \def\method(#1,#2,#3,#4,#5){
		            \coordinate (c#3) at ($(c#3)-(0,.4)$);
		            \coordinate (e#1) at ($(cmin)+(cmax)-(\s#2,0)$);
		            \draw[<-,#4,thick] (e#1) |- (c#3) node [black,#3=.1cm,inner sep=0,outer sep=0] {$\lambda = $ #5};
		        }
		    
		        \def\sigbar(#1,#2){
		            \coordinate (csig) at ($(csig) - (0,.2)$);    
		            \fill[black,rounded corners=.05cm] ($(e#1)-(.10,.05)+(csig)$) rectangle ($(e#2)+(.10,.05)+(csig)$);
		        }
		    
		        \method(CBND7,1.11,right,black,7)
		        \method(CBND5,2.00,right,black,5)
	
		        \method(CBND,4.00,left,black,1)	
		        \method(CBND3,2.88,left,black,3)
		
		        \sigbar(CBND3,CBND5);
		        \sigbar(CBND5,CBND7);
		    
		    \end{tikzpicture}
	    } \\
	    \vspace{0.5cm} \resizebox{0.3\textwidth}{!}
	    {
	        \begin{tikzpicture}[>=latex]
	
	            \def\nemnum{4}        
	            \def\signum{2}        
	            \def\s{1*}                
	            \def\cd{1.06}        
	
	            \coordinate (cmin) at (\s1,0);
	            \coordinate (cmax) at (\s\nemnum,0);
		        \coordinate (cleft) at ($(cmin)-(.2,(\signum*.25-.05)$);
		        \coordinate (cright) at ($(cmax)+(.2,-(\signum*.25-.05)$);
		        \coordinate (csig) at (0,-.15);
		    
		        \draw (cmin) -- (cmax);
		    
		        \foreach \c in {1,2,...,\nemnum}{
		            \draw ($(cmin)+(cmax)-(\s\c,0)$) -- ($(cmin)+(cmax)-(\s\c,-.2)$) node[above] {\c};
		             \ifthenelse{\c < \nemnum} {
		                \draw ($(\s\c,0)+(\s.5,0)$) -- ($(\s\c,0)+(\s.5,.1)$);
		              }
		        }
		    
		      
		        \def\method(#1,#2,#3,#4,#5){
		            \coordinate (c#3) at ($(c#3)-(0,.4)$);
		            \coordinate (e#1) at ($(cmin)+(cmax)-(\s#2,0)$);
		            \draw[<-,#4,thick] (e#1) |- (c#3) node [black,#3=.1cm,inner sep=0,outer sep=0] {$\lambda = $ #5};
		        }
		    
		        \def\sigbar(#1,#2){
		            \coordinate (csig) at ($(csig) - (0,.2)$);    
		            \fill[black,rounded corners=.05cm] ($(e#1)-(.10,.05)+(csig)$) rectangle ($(e#2)+(.10,.05)+(csig)$);
		        }
		    
		        \method(CBND5,1.85,right,black,5);
		        \method(CBND7,2.14,right,black,7);

		        \method(CBND,3.29,left,black,1)
		        \method(CBND3,2.71,left,black,3)

		        \sigbar(CBND3,CBND5);
		        \sigbar(CBND,CBND3);
		    
		    \end{tikzpicture}
	    }

    \caption{Critical differences charts for individual nested dichotomies. Top: Class balanced selection. Bottom: Random-pair selection.}
	\label{fig:cd_individual} 
\end{figure}

Figure~\ref{fig:cd_individual} shows critical difference plots for both subset selection methods. Class balanced selection shows a clear trend that increasing $\lambda$ improves the RMSE, with the average rank for $\lambda=1$ being exactly 4. For random-pair selection, choosing $\lambda=3$ is shown to be statistically equivalent to $\lambda=1$, while higher values of $\lambda$ give superior results on average.

\subsection{Ensembles of Nested Dichotomies}
Typically, nested dichotomies are utilised in an ensemble setting, so we investigate the predictive performance of ensembles of ten nested dichotomies with multiple subset evaluation, with bagging and AdaBoost employed as the ensemble methods.

\subsubsection*{Class Threshold.}
\begin{figure}[!t]
	\centering
		\begin{tabular}{l}
			\resizebox{0.45\textwidth}{!}
			{
			    \begin{tikzpicture}
			      \begin{axis} [
			        cycle list name=growthlist,
			        width=11cm,
			        height=7cm,
			        line width=1pt,                               
			        xmin=0.5,
			        xmax=7.5,
			        ymin=0.93,
			        ymax=1.22,
		            ytick={0.95, 1.0, 1.05, 1.1, 1.15, 1.2},
		            xtick={1,2,3,4,5,6,7},
		            xticklabels={,,},
			        grid=major,
			        ylabel={Relative RMSE (Bagging)},
			        enlarge y limits={abs=0},
			        legend style={
			          at={(0.04,0.95)}, anchor=north west
			        },
			      ]
			        \addplot+ [
			          error bars/.cd,
			            y dir=both, 
			            y explicit,
			            error bar style={line width=1pt},
		                error mark options={
					      rotate=90,
					      mark size=5pt,
					      line width=1pt
					    }
			        ]
			        coordinates {
			          (1,1)
			          (2,1)
			          (3,1.006) +- (0.013, 0.013)
			          (4,1.016) +- (0.026, 0.026)
			          (5,1.029) +- (0.043, 0.043)
			          (6,1.043) +- (0.058, 0.058)
				      (7,1.067) +- (0.092, 0.092)
			        };
	
					\addplot+ [
			          error bars/.cd,
			            y dir=both, 
			            y explicit,
			            error bar style={line width=1pt},
		                error mark options={
					      rotate=90,
					      mark size=5pt,
					      line width=1pt
					    }
			        ]
			        coordinates {
			          (1,1)
			          (2,1)
			          (3,1.008) +- (0.013, 0.013)
			          (4,1.016) +- (0.025, 0.025)
			          (5,1.044) +- (0.038, 0.038)
			          (6,1.051) +- (0.050, 0.050)
				      (7,1.078) +- (0.085, 0.085)
			        };
			        
			        \addlegendentry{Class Balanced}
			        \addlegendentry{Random-Pair}
			      \end{axis}
			    \end{tikzpicture}
		    } \\
		 \resizebox{0.45\textwidth}{!}
			{
			    \begin{tikzpicture}
			      \begin{axis} [
			        cycle list name=growthlist,
			        width=11cm,
			        height=7cm,
			        line width=1pt,                               
			        xmin=0.5,
			        xmax=7.5,
			        ymin=0.77,
			        ymax=1.33,
		            ytick={0.8, 0.9, 1.0, 1.1, 1.2, 1.3},
			        grid=major,
			        xlabel={Class Threshold},
			        ylabel={Relative RMSE (AdaBoost)},
			        enlarge y limits={abs=0},
			        legend style={
			          at={(0.04,0.95)}, anchor=north west
			        },
			      ]
			        \addplot+ [
			          error bars/.cd,
			            y dir=both, 
			            y explicit,
			            error bar style={line width=1pt},
		                error mark options={
					      rotate=90,
					      mark size=5pt,
					      line width=1pt
					    }
			        ]
			        coordinates {
			          (1,1)
			          (2,1)
			          (3,1.016) +- (0.047, 0.047)
			          (4,1.015) +- (0.048, 0.048)
			          (5,1.046) +- (0.132, 0.132)
				      (6,1.074) +- (0.233, 0.233)
  				      (7,1.073) +- (0.233, 0.233)
			        };
	
					\addplot+ [
			          error bars/.cd,
			            y dir=both, 
			            y explicit,
			            error bar style={line width=1pt},
		                error mark options={
					      rotate=90,
					      mark size=5pt,
					      line width=1pt
					    }
			        ]
			        coordinates {
			          (1,1)
			          (2,1)
			          (3,0.9999) +- (0.016, 0.016)
			          (4,1) +- (0.020, 0.020)
			          (5,1.0029) +- (0.025, 0.025)
			          (6,1.0078) +- (0.032, 0.032)
				      (7,1.0185) +- (0.046, 0.046)
			        };

			        \addlegendentry{Class Balanced}
			        \addlegendentry{Random-Pair}
			      \end{axis}
			    \end{tikzpicture}
		    }
		\end{tabular}
	\caption{\label{fig:threshold} Effect of changing the class threshold on RMSE for ensembles of nested dichotomies.}	
\end{figure}
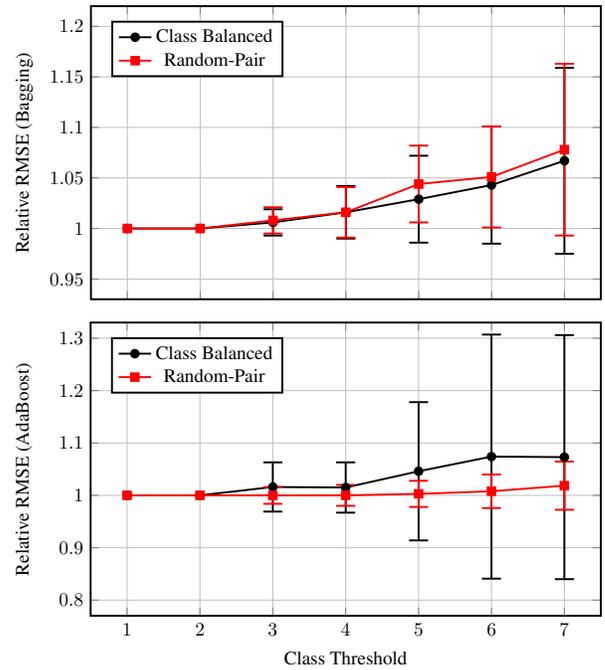

As previously discussed, the number of binary problems is reduced when multiple subset evaluation is applied. This could have negative a effect on ensemble diversity, and therefore potentially reduce predictive performance. To investigate this effect, we built ensembles of nested dichotomies with multiple subset evaluation by introducing a \textit{class threshold}, the number of classes present at a node required to perform multiple subset evaluation, and varying its value from one to seven. We plot the test RMSE, relative to having a class threshold of one, averaged over the datasets from Table~\ref{tab:datasets}, including standard errors, in Figure~\ref{fig:threshold}. Surprisingly, the RMSE increases monotonically, showing that the potentially reduced ensemble diversity does not have a negative effect on the RMSE for ensembles of this size. Therefore, we use a class threshold of one in our subsequent experiments. However, note that increasing the class threshold has a positive effect on training time, so it may be useful to apply it in practice.

\subsubsection*{Number of Subsets.}
\begin{figure}[t]
    \centering
    {
	    \resizebox{ 0.3\textwidth}{!}
	    {
	        \begin{tikzpicture}[>=latex]
	
	            \def\nemnum{4}        
	            \def\signum{2}        
	            \def\s{1*}                
	            \def\cd{1.06}        
	
	            \coordinate (cmin) at (\s1,0);
	            \coordinate (cmax) at (\s\nemnum,0);
		        \coordinate (cleft) at ($(cmin)-(.2,(\signum*.25-.05)$);
		        \coordinate (cright) at ($(cmax)+(.2,-(\signum*.25-.05)$);
		        \coordinate (csig) at (0,-.15);
		    
		        \draw (cmin) -- (cmax);
		    
		        \foreach \c in {1,2,...,\nemnum}{
		            \draw ($(cmin)+(cmax)-(\s\c,0)$) -- ($(cmin)+(cmax)-(\s\c,-.2)$) node[above] {\c};
		             \ifthenelse{\c < \nemnum} {
		                \draw ($(\s\c,0)+(\s.5,0)$) -- ($(\s\c,0)+(\s.5,.1)$);
		              }
		        }
		    
		        \draw[<->] ($(cmax)+(0,.7)$) -- node[above=.10cm,inner sep=0,outer sep=0] {\small{\textit{CD = 1.06}}} ($(cmax)-(\s\cd,-.7)$);
		      
		        \def\method(#1,#2,#3,#4,#5){
		            \coordinate (c#3) at ($(c#3)-(0,.4)$);
		            \coordinate (e#1) at ($(cmin)+(cmax)-(\s#2,0)$);
		            \draw[<-,#4,thick] (e#1) |- (c#3) node [black,#3=.1cm,inner sep=0,outer sep=0] {$\lambda = $ #5};
		        }
		    
		        \def\sigbar(#1,#2){
		            \coordinate (csig) at ($(csig) - (0,.2)$);    
		            \fill[black,rounded corners=.05cm] ($(e#1)-(.10,.05)+(csig)$) rectangle ($(e#2)+(.10,.05)+(csig)$);
		        }
		    
		        \method(CBND7,1.0,right,black,7)
		        \method(CBND5,1.94,right,black,5)
	
		        \method(CBND,4.00,left,black,1)	
		        \method(CBND3,2.94,left,black,3)
		
		        \sigbar(CBND5,CBND7);
		        \sigbar(CBND3,CBND5);
		    
		    \end{tikzpicture}
	    } \\
	    \vspace{0.5cm} \resizebox{0.3\textwidth}{!}
	    {
	        \begin{tikzpicture}[>=latex]
	
	            \def\nemnum{4}        
	            \def\signum{2}        
	            \def\s{1*}                
	            \def\cd{1.06}        
	
	            \coordinate (cmin) at (\s1,0);
	            \coordinate (cmax) at (\s\nemnum,0);
		        \coordinate (cleft) at ($(cmin)-(.2,(\signum*.25-.05)$);
		        \coordinate (cright) at ($(cmax)+(.2,-(\signum*.25-.05)$);
		        \coordinate (csig) at (0,-.15);
		    
		        \draw (cmin) -- (cmax);
		    
		        \foreach \c in {1,2,...,\nemnum}{
		            \draw ($(cmin)+(cmax)-(\s\c,0)$) -- ($(cmin)+(cmax)-(\s\c,-.2)$) node[above] {\c};
		             \ifthenelse{\c < \nemnum} {
		                \draw ($(\s\c,0)+(\s.5,0)$) -- ($(\s\c,0)+(\s.5,.1)$);
		              }
		        }
		    
		      
		        \def\method(#1,#2,#3,#4,#5){
		            \coordinate (c#3) at ($(c#3)-(0,.4)$);
		            \coordinate (e#1) at ($(cmin)+(cmax)-(\s#2,0)$);
		            \draw[<-,#4,thick] (e#1) |- (c#3) node [black,#3=.1cm,inner sep=0,outer sep=0] {$\lambda = $ #5};
		        }
		    
		        \def\sigbar(#1,#2){
		            \coordinate (csig) at ($(csig) - (0,.2)$);    
		            \fill[black,rounded corners=.05cm] ($(e#1)-(.10,.05)+(csig)$) rectangle ($(e#2)+(.10,.05)+(csig)$);
		        }
		    

		        \method(CBND5,2.00,right,black,5)
		        \method(CBND3,2.05,right,black,3)
	
		        \method(CBND,3.53,left,black,1)
		        \method(CBND7,2.41,left,black,7)

		        \sigbar(CBND7,CBND3);
		    
		    \end{tikzpicture}
	    }
    }

\caption{Critical differences charts for ensemble of ten bagged nested dichotomies. Top: Class balanced selection. Bottom: Random-pair selection.}
	\label{fig:cd_bagging} 
\end{figure}
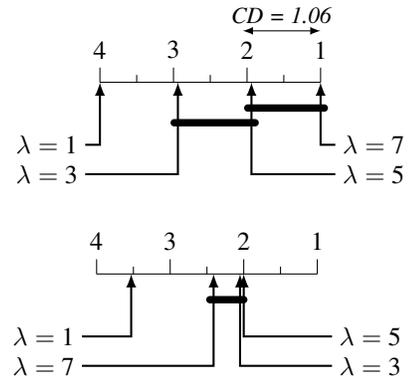

We now investigate the effect of $\lambda$ when using bagging and boosting. Figure~\ref{fig:cd_bagging} shows critical difference plots for bagging. Both subset selection methods improve when utilising multiple subset selection. In the case when class-balanced selection is used, as was observed for single nested dichotomies, the average ranks across all datasets closely correspond to the integer values, showing that increasing the number of subsets evaluated consistently improves performance. For random-pair selection, a more constrained subset selection method, each value of $\lambda > 1$ is statistically equivalent and superior to the single subset case.

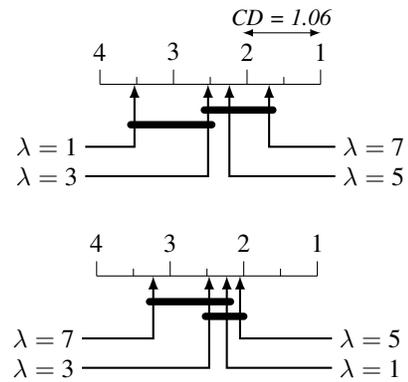
\begin{figure}[t]
    \centering
    {	    
	    \resizebox{ 0.3\textwidth}{!}
	    {
	        \begin{tikzpicture}[>=latex]
	
	            \def\nemnum{4}        
	            \def\signum{2}        
	            \def\s{1*}                
	            \def\cd{1.06}        
	
	            \coordinate (cmin) at (\s1,0);
	            \coordinate (cmax) at (\s\nemnum,0);
		        \coordinate (cleft) at ($(cmin)-(.2,(\signum*.25-.05)$);
		        \coordinate (cright) at ($(cmax)+(.2,-(\signum*.25-.05)$);
		        \coordinate (csig) at (0,-.15);
		    
		        \draw (cmin) -- (cmax);
		    
		        \foreach \c in {1,2,...,\nemnum}{
		            \draw ($(cmin)+(cmax)-(\s\c,0)$) -- ($(cmin)+(cmax)-(\s\c,-.2)$) node[above] {\c};
		             \ifthenelse{\c < \nemnum} {
		                \draw ($(\s\c,0)+(\s.5,0)$) -- ($(\s\c,0)+(\s.5,.1)$);
		              }
		        }
		    
		        \draw[<->] ($(cmax)+(0,.7)$) -- node[above=.10cm,inner sep=0,outer sep=0] {\small{\textit{CD = 1.06}}} ($(cmax)-(\s\cd,-.7)$);
		      
		        \def\method(#1,#2,#3,#4,#5){
		            \coordinate (c#3) at ($(c#3)-(0,.4)$);
		            \coordinate (e#1) at ($(cmin)+(cmax)-(\s#2,0)$);
		            \draw[<-,#4,thick] (e#1) |- (c#3) node [black,#3=.1cm,inner sep=0,outer sep=0] {$\lambda = $ #5};
		        }
		    
		        \def\sigbar(#1,#2){
		            \coordinate (csig) at ($(csig) - (0,.2)$);    
		            \fill[black,rounded corners=.05cm] ($(e#1)-(.10,.05)+(csig)$) rectangle ($(e#2)+(.10,.05)+(csig)$);
		        }
		    
		        \method(CBND7,1.70,right,black,7)
		        \method(CBND5,2.24,right,black,5)
	
		        \method(CBND,3.53,left,black,1)
		        \method(CBND3,2.53,left,black,3)

		        \sigbar(CBND3,CBND7);
		        \sigbar(CBND,CBND3);
		    
		    \end{tikzpicture}
	    } \\
	    \vspace{0.5cm} \resizebox{ 0.3\textwidth}{!}
	    {
	        \begin{tikzpicture}[>=latex]
	
	            \def\nemnum{4}        
	            \def\signum{2}        
	            \def\s{1*}                
	            \def\cd{1.06}        
	
	            \coordinate (cmin) at (\s1,0);
	            \coordinate (cmax) at (\s\nemnum,0);
		        \coordinate (cleft) at ($(cmin)-(.2,(\signum*.25-.05)$);
		        \coordinate (cright) at ($(cmax)+(.2,-(\signum*.25-.05)$);
		        \coordinate (csig) at (0,-.15);
		    
		        \draw (cmin) -- (cmax);
		    
		        \foreach \c in {1,2,...,\nemnum}{
		            \draw ($(cmin)+(cmax)-(\s\c,0)$) -- ($(cmin)+(cmax)-(\s\c,-.2)$) node[above] {\c};
		             \ifthenelse{\c < \nemnum} {
		                \draw ($(\s\c,0)+(\s.5,0)$) -- ($(\s\c,0)+(\s.5,.1)$);
		              }
		        }
		    
		      
		        \def\method(#1,#2,#3,#4,#5){
		            \coordinate (c#3) at ($(c#3)-(0,.4)$);
		            \coordinate (e#1) at ($(cmin)+(cmax)-(\s#2,0)$);
		            \draw[<-,#4,thick] (e#1) |- (c#3) node [black,#3=.1cm,inner sep=0,outer sep=0] {$\lambda = $ #5};
		        }
		    
		        \def\sigbar(#1,#2){
		            \coordinate (csig) at ($(csig) - (0,.2)$);    
		            \fill[black,rounded corners=.05cm] ($(e#1)-(.10,.05)+(csig)$) rectangle ($(e#2)+(.10,.05)+(csig)$);
		        }
		    

		        \method(CBND7,3.23,left,black,7)
				\method(CBND3,2.47,left,black,3)

		        \method(CBND5,2.05,right,black,5)		
		        \method(CBND,2.23,right,black,1)

		        \sigbar(CBND7,CBND);
		        \sigbar(CBND3,CBND5);
		    
		    \end{tikzpicture}
	    } 
	}
\caption{Critical differences charts for ensemble of ten nested dichotomies, ensembled with AdaBoost. Top: Class balanced selection. Bottom: Random-pair selection.}
	\label{fig:cd_boosting} 
\end{figure}

The critical difference plots in Figure~\ref{fig:cd_boosting} (top) show boosted nested dichotomies are significantly improved by increasing the number of subsets sufficiently when class-balanced nested dichotomies are used. Results are less consistent for random-pair selection, with few significant results in either direction. This is reflected in the critical differences plot (Fig.~\ref{fig:cd_boosting}, bottom), which shows single subset evaluation statistically equivalent to multiple subset selection for all values of $\lambda$, with $\lambda = 7$ performing markedly worse on average. As RMSE is based on probability estimates, this may be in part due to poor probability calibration, which is known to affect boosted ensembles~\cite{niculescu2005predicting} and nested dichotomies~\cite{leathart2018calibration}.

\section{Conclusion\label{sec:conclusion}}
Multiple subset selection in nested dichotomies can improve predictive performance while retaining the particular advantages of the subset selection method employed. We present an analysis of the effect of multiple subset selection on expected RMSE and show empirically in our experiments that adopting our technique can improve predictive performance, at the cost of a constant factor in training time. 

The results of our experiments suggest that for class-balanced selection, performance can be consistently improved significantly by utilising multiple subset evaluation. For random-pair selection, $\lambda=3$ yields the best trade-off between predictive performance and training time, but when AdaBoost is used, our experiments show that multiple subset evaluation is not generally beneficial.

Avenues of future research include comparing multiple subset evaluation with base learners other than logistic regression. It is unlikely that training RMSE of the internal models will be a reliable indicator when selecting splits based on more complex models such as decision trees or random forests, so other metrics may be needed. Also, it may be beneficial to choose subsets such that maximum ensemble diversity is achieved, possibly through information theoretic measures such as variation of information~\cite{meilua2003comparing}. Existing meta-heuristic approaches to constructing individual nested dichotomies like genetic algorithms~\cite{lee2003binary,wever2018ensembles} could also be adapted to optimise ensembles in this way.

\section{Acknowledgements}
This research was supported by the Marsden Fund Council from Government funding, administered by the Royal Society of New Zealand.

\bibliographystyle{aaai}
\bibliography{multi_subset_nd}

\end{document}